\documentclass[conference]{IEEEtran}
\IEEEoverridecommandlockouts
\usepackage{cite}
\usepackage{amsmath,amssymb,amsfonts}
\usepackage{algorithmic}
\usepackage{graphicx}
\usepackage{textcomp}
\usepackage{xcolor}
\usepackage{booktabs}
\usepackage{makecell}
\usepackage{multirow}

\def\BibTeX{{\rm B\kern-.05em{\sc i\kern-.025em b}\kern-.08em
    T\kern-.1667em\lower.7ex\hbox{E}\kern-.125emX}}
\begin{document}

\title{Comparing LSTM-Based Sequence-to-Sequence Forecasting Strategies for 24-Hour Solar Proton Flux Profiles Using GOES Data\\
\thanks{}
}

\author{
\IEEEauthorblockN{Kangwoo Yi}
\IEEEauthorblockA{\textit{New Jersey Institute of Technology} \\
Newark, USA \\
kangwoo.yi@njit.edu}
\and
\IEEEauthorblockN{Bo Shen*\thanks{*Dr. Bo Shen is the corresponding author.}}
\IEEEauthorblockA{\textit{New Jersey Institute of Technology} \\
Newark, USA \\
bo.shen@njit.edu}
\and
\IEEEauthorblockN{Qin Li}
\IEEEauthorblockA{\textit{New Jersey Institute of Technology} \\
Newark, USA \\
ql47@njit.edu}
\and
\IEEEauthorblockN{Haimin Wang}
\IEEEauthorblockA{\textit{NJIT} \\
Newark, USA \\
haimin.wang@njit.edu}
\and
\IEEEauthorblockN{Yong-Jae Moon}
\IEEEauthorblockA{\textit{Kyung Hee University} \\
Yongin, Republic of Korea \\
moonyj@khu.ac.kr}
\and
\IEEEauthorblockN{Jaewon Lee}
\IEEEauthorblockA{\textit{Kyung Hee University} \\
Yongin, Republic of Korea \\
jaewonlee@khu.ac.kr}
\and
\IEEEauthorblockN{Hwanhee Lee}
\IEEEauthorblockA{\textit{Korea Astronomy and Space Science Institute} \\
Daejeon, Republic of Korea \\
hhee@kasi.re.kr}
}
\maketitle

\begin{abstract}
Solar Proton Events (SPEs) cause significant radiation hazards to satellites, astronauts, and technological systems. Accurate forecasting of their proton flux time profiles is crucial for early warnings and mitigation. This paper explores deep learning sequence-to-sequence (seq2seq) models based on Long Short-Term Memory networks to predict 24-hour proton flux profiles following SPE onsets. We used a dataset of 40 well-connected SPEs (1997–2017) observed by NOAA GOES, each associated with a $\geq$M-class western-hemisphere solar flare and undisturbed proton flux profiles. Using 4-fold stratified cross-validation, we evaluate seq2seq model configurations (varying hidden units and embedding dimensions) under multiple forecasting scenarios: (i) proton-only input vs. combined proton+X-ray input, (ii) original flux data vs. trend-smoothed data, and (iii) autoregressive vs. one-shot forecasting. 
Our major results are as follows: 
First, one-shot forecasting consistently yields lower error than autoregressive prediction, avoiding the error accumulation seen in iterative approaches. 
Second, on the original data, proton-only models outperform proton+X-ray models. However, with trend-smoothed data, this gap narrows or reverses in proton+X-ray models.
Third, trend-smoothing significantly enhances the performance of proton+X-ray models by mitigating fluctuations in the X-ray channel.
Fourth, while models trained on trend-smoothed data perform best on average, the best-performing model was trained on original data, suggesting that architectural choices can sometimes outweigh the benefits of data preprocessing.
\end{abstract}

\begin{IEEEkeywords}
Solar proton event, Time series, Deep Learning
\end{IEEEkeywords}

\section{INTRODUCTION}
Solar proton events (SPEs) are bursts of high-energy particles in interplanetary space accelerated by solar flares and coronal mass ejections (CMEs) \cite{b1}, \cite{b2}. 
When these particles reach near-Earth space, they can cause spacecraft single event upset, communication blackouts in polar regions, and elevated radiation exposure to astronauts and high-altitude flight. 
Given these impacts, forecasting SPE occurrence and intensity is a challenge in space weather research. The National Oceanic and Atmospheric Administration (NOAA) defines a solar proton event as an event that exceeds 10 pfu (particle flux unit) of $\geq$10 MeV protons, and warns them with the S scale in NOAA space weather scales.

Previous studies on SPE prediction have primarily focused on forecasting the occurrence, onset time, or peak flux of SPEs using solar flare, CME, or radio burst data \cite{b3, b4, b5, b6, b7, b8}. While physics-based models simulate particle acceleration and transport \cite{b9, b10}, and empirical models forecast SPE flux profiles \cite{b11}, both are limited in real-time applications due to their computational intensity and/or the lack of timely data availability.

Beyond predicting if an event will occur, forecasting the time profile of proton flux (i.e., how the solar proton flux evolves over hours to days) is crucial for assessing the duration and peak of exposure, as SPEs can last from hours to weeks. However, forecasting the time series of SPE flux profiles is challenging due to several factors: 

\textbullet\ Rarity of events: SPEs above the NOAA S1 level ($\geq$10 pfu) occur infrequently, providing limited historical samples for training and pattern recognition. According to the NOAA SPE event list, only 134 events were recorded from 1997 to 2017.

\textbullet\ Impulsive, non-repetitive profiles: Each event’s proton flux can spike by orders of magnitude in a short time and then decay nonlinearly, with large variations in peak intensity and duration, and no seasonal or periodic patterns to leverage.

\textbullet\ Multi-input variability: The flux profile depends on complex solar eruption parameters (flare intensity, CME speed, magnetic connectivity), which are hard to quantify for real-time modeling.

These challenges mean traditional time series forecasting techniques struggle with SPE profiles, and physics-based models require detailed eruption inputs that may not be available in real-time. 

In this paper, we compare a range of LSTM seq2seq architectures and forecasting strategies for near-real-time solar proton flux profile prediction in SPE situations. We construct a carefully filtered dataset of 40 historical SPE events, each characterized by well-connected solar conditions, and evaluate performance via cross-validation. Our contributions include: (1) demonstrating a seq2seq LSTM approach for 24-hour proton flux profile forecasting, (2) systematically assessing the impact of including solar X-ray data as an additional input, of trend-smoothed data preprocessing, and of autoregressive vs. one-shot prediction strategies, and (3) providing insights into model configuration choices (hidden layer size and embedding dimensions) on a small but domain-specific dataset. The following sections describe the dataset and preprocessing, the LSTM seq2seq methodology and experimental setup, results with a comparative analysis of strategies, discussion of implications, and conclusions.

\section{RELATED WORK}
The prediction of solar energetic particle (SEP) events, particularly in the context of space weather forecasting, has been approached through both empirical and physics-based methodologies. Early efforts have primarily focused on establishing relationships between observable solar phenomena and subsequent SEP signatures at 1 AU.

Luhmann et al. \cite{b10} modeled the 12 May 1997 SEP event using a forward simulation based on a heliospheric MHD model from Odstrcil et al. \cite{b12}. The shock and magnetic field structure from the simulation were used to trace field-line connections to the observer, assuming scatter-free transport.

Early forecasting methods for solar energetic particle events often relied on empirical correlations with solar flare and CME observations. Ji et al. \cite{b11} developed a model to predict $\geq$ 10 MeV proton flux profiles for well-connected SPEs by fitting each event’s flux curve with a modified Weibull function and linking the function parameters (peak flux, rise time, decay time) to the associated soft X-ray flare measurements. Using 49 SPEs with flare data, they reported a correlation of 0.65, which improved to 0.83 with CME inputs, based on 22 SPEs with available CME information.

In recent years, deep learning has offered new opportunities for time series forecasting in space weather tasks. For example, Yi et al. \cite{b13} applied a LSTM sequence-to-sequence (seq2seq) model with attention to forecast solar X-ray flare flux profiles, achieving significant improvements over conventional regression models. Their model was trained on GOES soft X-ray (0.1–0.8 nm) data and could predict the flare’s 30-minute rise-phase profile, even for the most impulsive large flares. This illustrates that seq2seq networks can capture rapid, non-linear surges in solar flux data. This success with flare time series suggests similar architectures could learn the temporal patterns of proton flux surges in SPEs.

LSTM-based sequence-to-sequence \cite{b14} has gained widespread adoption in various time series applications due to their ability to encode complex temporal dependencies \cite{b15, b16, b17}. Seq2seq structure typically consists of an encoder LSTM that processes the input sequence and a decoder LSTM that generates the output sequence. In space weather contexts, seq2seq models have shown potential for capturing the steep rises and gradual decay patterns characteristic of solar flares and proton events. However, there is limited systematic evaluation of different forecasting schemes (e.g., autoregressive vs. one-shot) and model hyperparameters (e.g., hidden size, embedding dimension) for SPE flux profile prediction, which this study aims to address.

\section{Dataset}

We assembled a dataset of 40 well-connected historical SPEs from 1997 to 2017 using NOAA GOES satellite measurements (i.e., the solar source is magnetically connected to Earth for efficient particle transport). Because SPEs are rare, each individual event can significantly affect model performance. To avoid performance degradation from noisy or low-quality events, we applied strict selection criteria :

\textbullet\ Each event is associated with a $\geq$M-class solar flare located in the western hemisphere of the Sun.

\textbullet\ The proton flux increase begins within ~4 hours after the flare peak, indicating a likely causal relationship

\textbullet\ Events with proton flux profiles that are severely disturbed or involve multiple overlapping SPEs are excluded. 

\begin{figure}[htbp]
\centerline{\includegraphics[width=\linewidth]{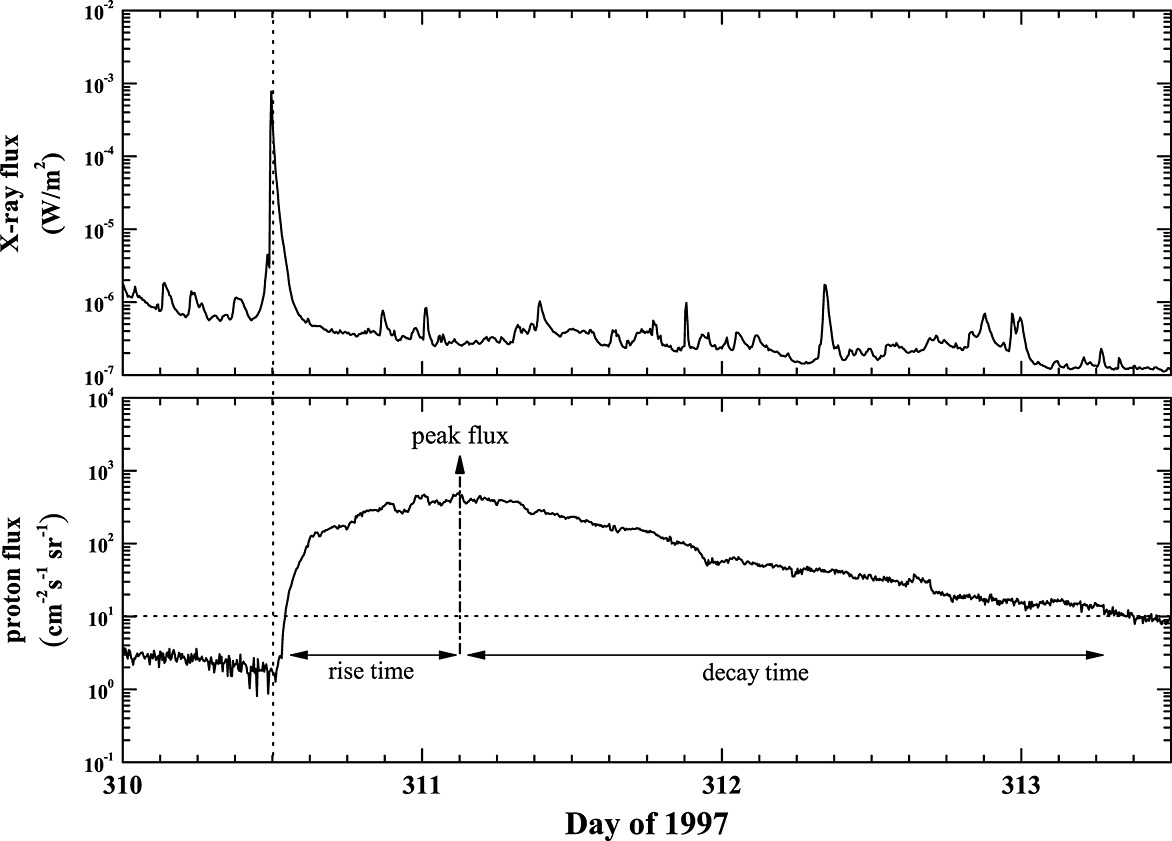}}
\caption{Example of well-connected SPE on 6 November 1997, from \cite{b11}. (top) GOES X-ray flux and (bottom) solar proton flux data. Vertical dashed line indicates flare peak time. Horizontal dashed line indicates threshold of SPE (10 pfu). }
\label{fig:SPE_example}
\end{figure}

Figure~\ref{fig:SPE_example} shows an example of a well-connected SPE with rise and decay times. As shown in Figure~\ref{fig:SPE_example}, the proton flux starts to increase after the flare peak time and has a peak flux after 18 h.

These criteria yielded a focused set of 40 SPEs that are strong enough and reasonably isolated for profile modeling. The events were further categorized by peak intensity into NOAA S-scale classes S1 (20 events), S2 (12 events), S3 (6 events), and S4 (2 events). Given the limited number of events (40), we employ 4-fold cross-validation to maximize training data while obtaining robust performance estimates. The folds are stratified such that each fold has a similar proportion of S1, S2, and S3/S4 events (S4 events merged with S3 due to low counts). In each cross-validation run, 75\% of events (30 events) are used for training, 25\% (10 events) for testing. The performance metrics reported are averaged over the 4 test folds. Table ~\ref{tab:dataset} shows the summary of the SPE dataset.

\begin{table}[htbp]
\caption{Well-connected SPE dataset (4-fold cross-validation)}
\begin{center}
\begin{tabular}{lcccc}
\hline
           & S1 & S2 & S3+S4 & Total \\
\hline
Training   & 15 & 9  & 6     & 30    \\
Test       & 5  & 3  & 2     & 10    \\
\hline
\end{tabular}
\label{tab:dataset}
\end{center}
\end{table}

For each event, we extracted the GOES $\geq$ 10 MeV proton flux time series from the event onset (when the $\geq$ 10 MeV proton flux first exceeds 10 pfu) to the event end (when it drops below 10 pfu). Using a sliding time-window approach, we constructed sequences centered on each time step in the event time series, covering 24 hours before and after the current point. Given a temporal resolution of 5 minutes, each sequence consists of 288 input points (past 24 hours) and 288 target points (future 24 hours) for model training.
We applied the same sequence generation method to the GOES 0.1–0.8nm X-ray flux data over the same time range as the corresponding proton flux data. Following the GOES X-ray data guidelines, the flux values were divided by 0.7. To align the dynamic range with that of the proton flux data, we additionally multiplied the values by 1E7 as a custom normalization step.

Each event time series was prepared in two formats for experiments: (a) the original proton flux, and (b) a trend-smoothed version of the proton flux (and X-ray flux, when used) obtained by applying a 1-hour sliding average. The smoothing was done with a ±30 min window around each point, which preserves the overall rise-fall trend but filters out high-frequency fluctuations. In both formats, flux values were log-transformed to reduce dynamic range and stabilize variance, which is standard for handling the highly skewed distributions typical of solar particle flux data.

\section{Methodology}
\subsection{Model variation}
We employ a seq2seq model to forecast the 24-hour proton flux profile from the preceding 24-hour data window. The seq2seq architecture consists of an encoder and a decoder, each composed of two stacked LSTM layers. The encoder processes the 24h (288 time steps) input sequence and compresses the information into a fixed-length internal state vector. The final hidden state of the encoder is passed through a dense layer to produce a lower-dimensional embedding vector. This embedding, together with the previously predicted proton flux, is used as input to an attention module that generates a context vector. The decoder then uses this attention-informed context, along with the embedding, to initialize its hidden state and generate the 24h (288 time steps) forecast sequence.

We systematically varied two key architecture hyperparameters: the number of LSTM units (hidden dimensionality in each LSTM layer) and the embedding size. We tried LSTM hidden unit sizes of 1024, 768, and 512, in combination with embedding vector sizes of 20, 16, 8, 4, and 1. This yielded 3×5 = 15 distinct model architectures to compare, ranging from a large model with 1024 units and a rich 20-dimensional embedding, down to a highly compressed model with 512 units and a 1-dimensional embedding (essentially forcing the entire profile’s state into a single scalar). All models were trained using the same data splits and settings to enable fair comparison. We used the mean squared error loss function in log flux units and optimized with the Adam optimizer with a 0.001 learning rate. 

\subsection{Forecasting strategies}
We evaluate six forecasting strategies arising from combinations of input data, data processing, and prediction mode:

\textbullet\ Input features: We compare a mono-feature input (using only the proton flux time series) versus a multi-input (proton+X-ray). In the multi-input case, the encoder LSTM takes proton and X-ray data. The decoder in both cases outputs only the proton flux profile. This tests whether adding real-time flare data improves the proton forecast.

\textbullet\ Preprocessing method: We evaluate models on original vs. trend-smoothed versions of the input data. The original data includes all the short-term variability, whereas the smoothed data emphasizes the 1-hour trend. This tests if the model benefits from a de-noised input. We apply the same smoothing to the X-ray when included.

\textbullet\ Forecasting mode: We implement the seq2seq decoder in two modes: autoregressive and one-shot forecasting. In the autoregressive mode, the decoder predicts the output sequence step-by-step, each time step output (e.g., 5 min interval) is fed back as input for predicting the next step. In the one-shot forecasting mode, the decoder produces the entire 24-hour sequence at once, without iterative feedback.

By combining the above options, we train and evaluate models under experimental settings: (mono vs. multi) × (original vs. trend) × {autoregressive vs. one-shot}. Trend-smoothed data is only used for the one-shot forecasting mode. All six strategies share the same architecture search space (the 15 LSTM unit/embedding configurations) and training procedure.

\subsection{Model performance evaluation}
For evaluation, we performed 4-fold stratified cross-validation on the 40 events, ensuring each fold had a similar mix of event intensities. Due to the small number of S4 events, we merged S3 and S4 into one category. In each fold, 75\% of the events were used for training and 25\% for testing, rotating such that every event appears in exactly one test fold. The two S4 events are assigned to different folds.

We evaluate model performance using root mean square error (RMSE), computed in logarithmic flux units across all time points. To provide a scale-normalized measure, we also calculate the percentage error. Performance metrics are then averaged over all folds. 

\begin{figure}[htbp]
\centering
\includegraphics[width=\linewidth]{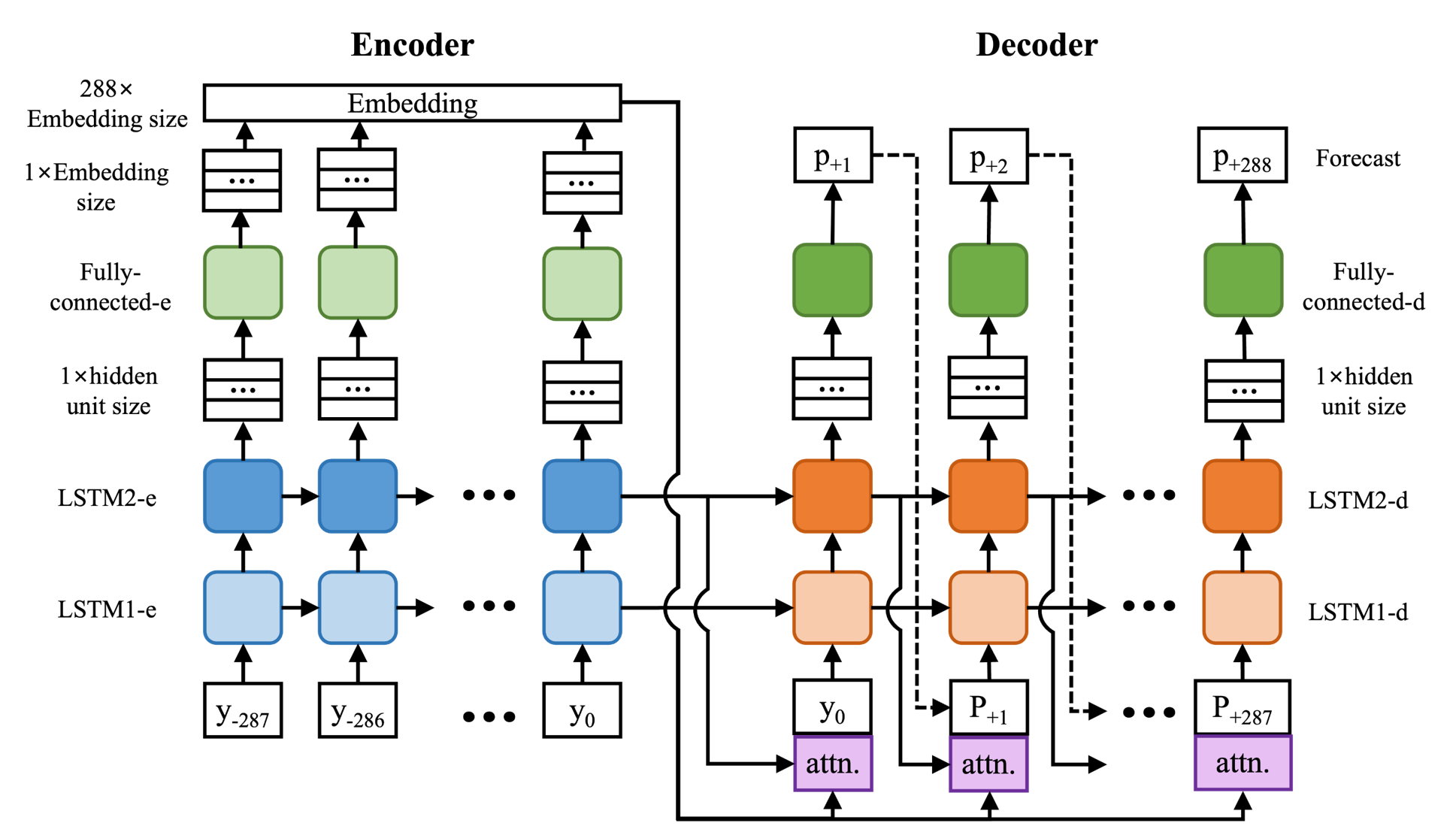}
\vspace{1mm}  
\includegraphics[width=\linewidth]{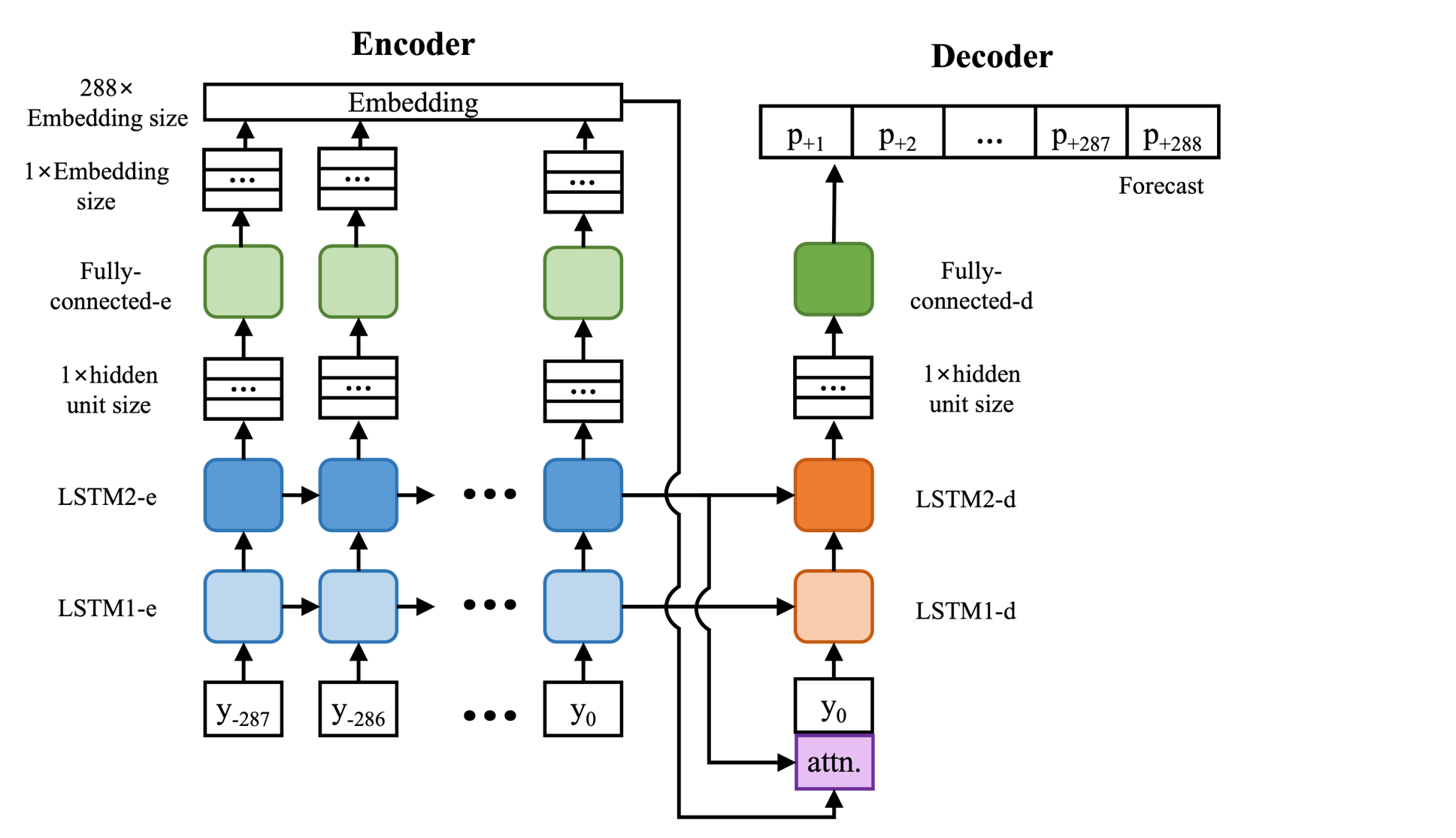}
\caption{Model structures. (top) Autoregressive forecasting mode and (bottom) one-shot forecasting mode. Blue and orange boxes represent LSTM layers in the encoder and decoder, respectively. Green boxes denote fully connected layers. White boxes represent input/output arrays or intermediate variables, and purple boxes indicate attention modules.}
\label{fig:model_structure}
\end{figure}

\section{EXPERIMENTS}
We evaluated 15 model structures across six forecasting strategies using 4-fold cross-validation. Table~\ref{tab:model_performances} summarizes the results, reporting the RMSE and the percentage error averaged over all folds. 
Each model name follows the format [Input]\_[DataType]\_[Forecasting], where:

\textbullet\ P refers to models using $\geq$ 10 MeV proton flux data only, while P+XR includes both the $\geq$ 10 MeV proton and the 0.1--0.8nm X-ray flux data.

\textbullet\ orig uses the original time series, while trend uses trend-smoothed data.

\textbullet\ AR (autoregressive) produces the forecast sequentially step-by-step, while OS (one-shot) outputs the entire forecast at once.

\begin{table*}[t]
\centering
\caption{Forecasting performance of different seq2seq configurations under multiple input and preprocessing settings. Metrics reported as RMSE // Percentage Error averaged over all folds. The models with RMSE lower than 0.310 are highlighted in bold.}
\label{tab:model_performances}
\renewcommand{\arraystretch}{1.3}
\begin{tabular}{lcccccc}
\toprule
\multirow{2}{*}{\textbf{\makecell{Model structure \\ (Unit--Embed Size)}}} 
& \multicolumn{4}{c}{\textbf{Original data}} 
& \multicolumn{2}{c}{\textbf{Trend-smoothed data}} \\
\cmidrule(lr){2-5} \cmidrule(lr){6-7}
& \textbf{P\_orig\_AR} & \textbf{P\_orig\_OS} 
& \textbf{P+XR\_orig\_AR} & \textbf{P+XR\_orig\_OS}
& \textbf{P\_trend\_OS} & \textbf{P+XR\_trend\_OS} \\
\midrule[0.75pt]
\vspace{1mm}
1024-20 & 0.389 // 15.68\% & 0.320 // 12.05\% & 0.400 // 15.73\% & 0.334 // 12.45\% & 0.326 // 11.63\% & 0.312 // 11.77\% \\
1024-16 & 0.361 // 13.20\% & 0.318 // 11.86\% & 0.431 // 17.43\% & 0.337 // 12.68\% & 0.318 // 11.98\% & 0.318 // 12.13\% \\
1024-8  & 0.345 // 13.01\% & 0.327 // 12.23\% & 0.369 // 13.50\% & 0.329 // 11.93\% & 0.320 // 11.86\% & \textbf{0.306 // 11.12\%} \\
1024-4  & 0.341 // 12.36\% & 0.321 // 11.56\% & 0.362 // 13.32\% & 0.335 // 12.69\% & \textbf{0.309 // 11.10\%} & 0.312 // 11.08\% \\
1024-1  & 0.322 // 11.83\% & 0.320 // 11.96\% & 0.329 // 12.82\% & 0.315 // 11.73\% & 0.313 // 11.28\% & \textbf{0.305 // 10.88\%} \\
\midrule[0.75pt]
\vspace{1mm}
768-20  & 0.403 // 15.89\% & 0.321 // 11.31\% & 0.396 // 14.96\% & 0.319 // 11.54\% & 0.338 // 12.45\% & 0.314 // 11.42\% \\
768-16  & 0.363 // 13.21\% & 0.316 // 11.52\% & 0.440 // 17.81\% & 0.338 // 13.10\% & 0.324 // 11.85\% & 0.311 // 11.59\% \\
768-8   & 0.382 // 14.45\% & 0.319 // 11.65\% & 0.369 // 13.36\% & 0.353 // 13.46\% & \textbf{0.303 // 11.25\%} & \textbf{0.308 // 11.58\%} \\
768-4   & 0.325 // 12.57\% & 0.314 // 11.96\% & 0.388 // 14.75\% & 0.319 // 11.43\% & 0.312 // 11.41\% & \textbf{0.306 // 11.23\%} \\
768-1   & 0.326 // 11.91\% & 0.321 // 12.54\% & 0.339 // 12.39\% & 0.316 // 11.68\% & 0.317 // 11.91\% & 0.312 // 11.41\% \\
\midrule[0.75pt]
\vspace{1mm}
512-20  & 0.370 // 13.53\% & 0.316 // 11.49\% & 0.440 // 16.47\% & 0.317 // 11.70\% & 0.315 // 11.84\% & 0.318 // 11.70\% \\
512-16  & 0.367 // 13.91\% & \textbf{0.309 // 11.64\%\textbf} & 0.410 // 15.76\% & 0.325 // 11.76\% & \textbf{0.310 // 11.95\%} & 0.320 // 11.81\% \\
512-8   & 0.351 // 13.43\% & \textbf{0.303 // 11.03\%} & 0.377 // 13.94\% & 0.319 // 12.38\% & 0.314 // 11.64\% & 0.313 // 11.50\% \\
512-4   & 0.342 // 12.93\% & 0.319 // 11.71\% & 0.375 // 14.46\% & 0.323 // 11.73\% & 0.315 // 11.98\% & 0.313 // 11.83\% \\
512-1   & 0.322 // 11.76\% & \textbf{0.310 // 11.17\%} & 0.359 // 13.96\% & 0.319 // 11.48\% & \textbf{0.306 // 11.12\%} & \textbf{0.305 // 11.32\%} \\
\bottomrule
\end{tabular}
\end{table*}

As Table~\ref{tab:model_performances} shows, there are clear patterns in the results:

\textbullet\ \textbf{One-shot vs. Autoregressive:} One-shot forecasting generally outperforms the autoregressive approach. For most model configurations, the one-shot model error is lower than the corresponding autoregressive model error when comparing P\_orig\_AR vs. P\_orig\_OS, and similarly for the proton+X-ray cases. For instance, using a 512-8 model, the RMSE drops from approximately 0.351 in autoregressive to 0.303 in one-shot. A similar but smaller gap is also observed in some larger models.

This result aligns with the theoretical expectation that iterative prediction accumulates error at each time step, degrading the quality forecasts. By contrast, one-shot models benefit from the encoder’s global context and generate the full output sequence in one pass, avoiding error propagation. In practical terms, autoregressive models often underestimate the latter part of the profile due to early-stage deviations, whereas one-shot models tend to capture the overall shape more consistently.

\textbullet\ \textbf{Proton-only vs. Proton+X-ray Input:}
Adding the 0.1--0.8~nm X-ray flux as an additional input did not improve forecasting performance when using the original (non-detrended) data. In both autoregressive and one-shot modes, proton-only models consistently outperformed the multi-input (proton+X-ray) models in this setting. For example, in the one-shot original data configuration, the best proton-only model (512-8) achieved RMSE 0.303 with approximately 11.02\% error, whereas the best proton+X-ray model (1024-1) yielded RMSE 0.315 and 11.74\% error. While the performance gap is modest, it is consistently in favor of proton-only models on raw input (see P\_orig\_OS vs. P+XR\_orig\_OS in Table~\ref{tab:model_performances}).

However, when trend-smoothed data are used, this gap narrows or even reverses in some configurations. In particular, for smaller model architectures, the multi-input (proton+X-ray) models occasionally outperform their proton-only counterparts. For example, in the one-shot trend data configuration, the best proton-only model (768-8) achieved RMSE 0.303 with approximately 11.25\% error, whereas the best proton+X-ray model (1024-1) yielded RMSE 0.305 and 10.88\% error. This suggests that the proton time series alone is generally more effective for modeling the post-onset evolution of SPE flux profiles. In original data, the X-ray signal tends to introduce sharp, high-frequency components that can mislead the model. Smoothing preprocess mitigates this issue by emphasizing the trend in the X-ray data.

\textbullet\ \textbf{Effect of Trend Processing:}
Applying sliding-window smoothing to the input data had little effect on the proton-only models but significantly improved performance in the proton+X-ray input setting. For proton-only models, performance on original and smoothed data was nearly identical, suggesting that the LSTM could internally learn the baseline trend. However, for proton+X-ray models, smoothing the input consistently reduced both RMSE and percentage error. For instance, the best proton+X-ray one-shot model with trend-smoothed data (1024-1) achieved RMSE 0.305, improving upon the best value from the original data (1024-1) case (0.315). This gain can be attributed to the fact that smoothing the input reduces high-frequency fluctuations in the X-ray flux. By averaging the input over a sliding window, the model receives a more stable representation of the flare’s energy release and its potential effect on proton flux evolution. Although the improvement is modest, it suggests that input smoothing can enhance the compatibility between heterogeneous inputs in multi-modal forecasting tasks. Nonetheless, even with smoothing, the multi-input models did not outperform the best proton-only models. This implies that the added X-ray information was either redundant for the forecasting task or not effectively utilized by our current model architecture.

Notably, this trend-smoothing advantage does not hold universally. The best-performing configuration in our experiments was a model trained on original data, which achieved the lowest RMSE among all tested settings. This suggests that while trend-smoothed data is generally favorable, original data may still be competitive or even superior in certain configurations, especially when paired with optimized architecture and training strategy. 

To evaluate the robustness of the results, we present fold-wise RMSEs and their standard deviations for the six best-performing model configurations in each forecasting strategy, as summarized in Table~\ref{tab:foldwise}. The one-shot approach shows more stable performance across folds than the autoregressive approach. For instance, P\_orig\_AR shows a much larger standard deviation in RMSE compared with P\_orig\_OS model, and P+XR\_orig\_AR shows a much larger standard deviation in percentage error compared with P+XR\_orig\_OS model.
Across all six models, substantial differences appear between CV3 and the other folds in RMSE, and between CV4 and the other folds in percentage error, indicating that model evaluation can be strongly influenced by the choice of test dataset. This highlights the importance of cross-validation as a robust evaluation method for forecasting models with small dataset. 
Furthermore, opposite trends between RMSE and percentage error, for example in CV3 and CV4 of P\_trend\_OS, demonstrate that reliance on a single evaluation metric may lead to biased interpretations.

\begin{table*}[t]
\centering
\caption{Fold-wise RMSE and percentage error (mean $\pm$ standard deviation) for the six best-performing model configurations in each forecasting strategy. CV1–CV4 indicate each of the four cross-validation folds.}
\label{tab:foldwise}
\begin{tabular}{lcccccc}
\toprule
\textbf{Forecasting strategy} & \textbf{Model structure} &
\makecell[c]{\textbf{RMSE // Percentage error}\\ \textbf{(Standard deviation)}} &
\textbf{CV1} & \textbf{CV2} & \textbf{CV3} & \textbf{CV4} \\
\midrule
P\_orig\_AR       & 512-1   & \makecell[c]{0.322 // 11.76\%\\(0.027 // 1.22)} & 0.289 // 10.10\% & 0.313 // 12.37\% & 0.363 // 11.22\% & 0.323 // 13.34\% \\
\midrule
P\_orig\_OS       & 512-8   & \makecell[c]{0.303 // 11.03\%\\(0.003 // 1.24)} & 0.303 // 10.13\% & 0.298 // 11.06\% & 0.307 //  9.88\% & 0.304 // 13.03\% \\
\midrule
P+XR\_orig\_AR   & 1024-1  & \makecell[c]{0.329 // 12.82\%\\(0.030 // 1.54)} & 0.282 // 10.62\% & 0.333 // 13.18\% & 0.365 // 12.53\% & 0.336 // 14.94\% \\
\midrule
P+XR\_orig\_OS   & 1024-1  & \makecell[c]{0.315 // 11.73\%\\(0.026 // 0.99)} & 0.290 // 10.25\% & 0.302 // 11.43\% & 0.359 // 12.56\% & 0.309 // 12.69\% \\
\midrule
P\_trend\_OS      & 768-8   & \makecell[c]{0.303 // 11.25\%\\(0.016 // 0.93)} & 0.288 // 10.22\% & 0.293 // 11.16\% & 0.330 // 10.86\% & 0.301 // 12.75\% \\
\midrule
P+XR\_trend\_OS  & 1024-1  & \makecell[c]{0.305 // 10.88\%\\(0.014 // 1.04)} & 0.296 //  9.78\% & 0.314 // 11.76\% & 0.321 //  9.91\% & 0.287 // 12.07\% \\
\bottomrule
\end{tabular}
\end{table*}

\begin{figure}[htbp]
\centering
\includegraphics[width=\linewidth]{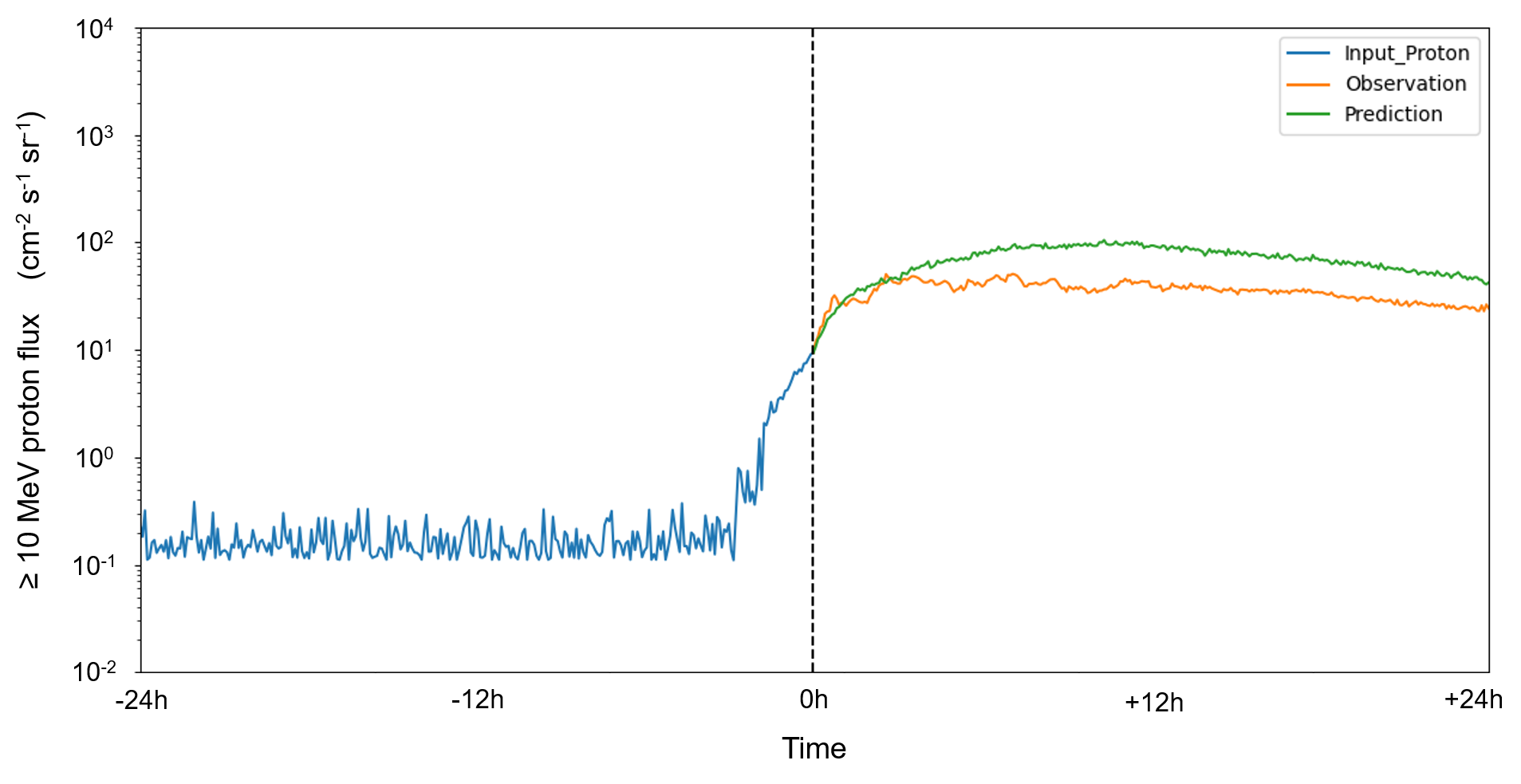}
\vspace{1mm}
\includegraphics[width=\linewidth]{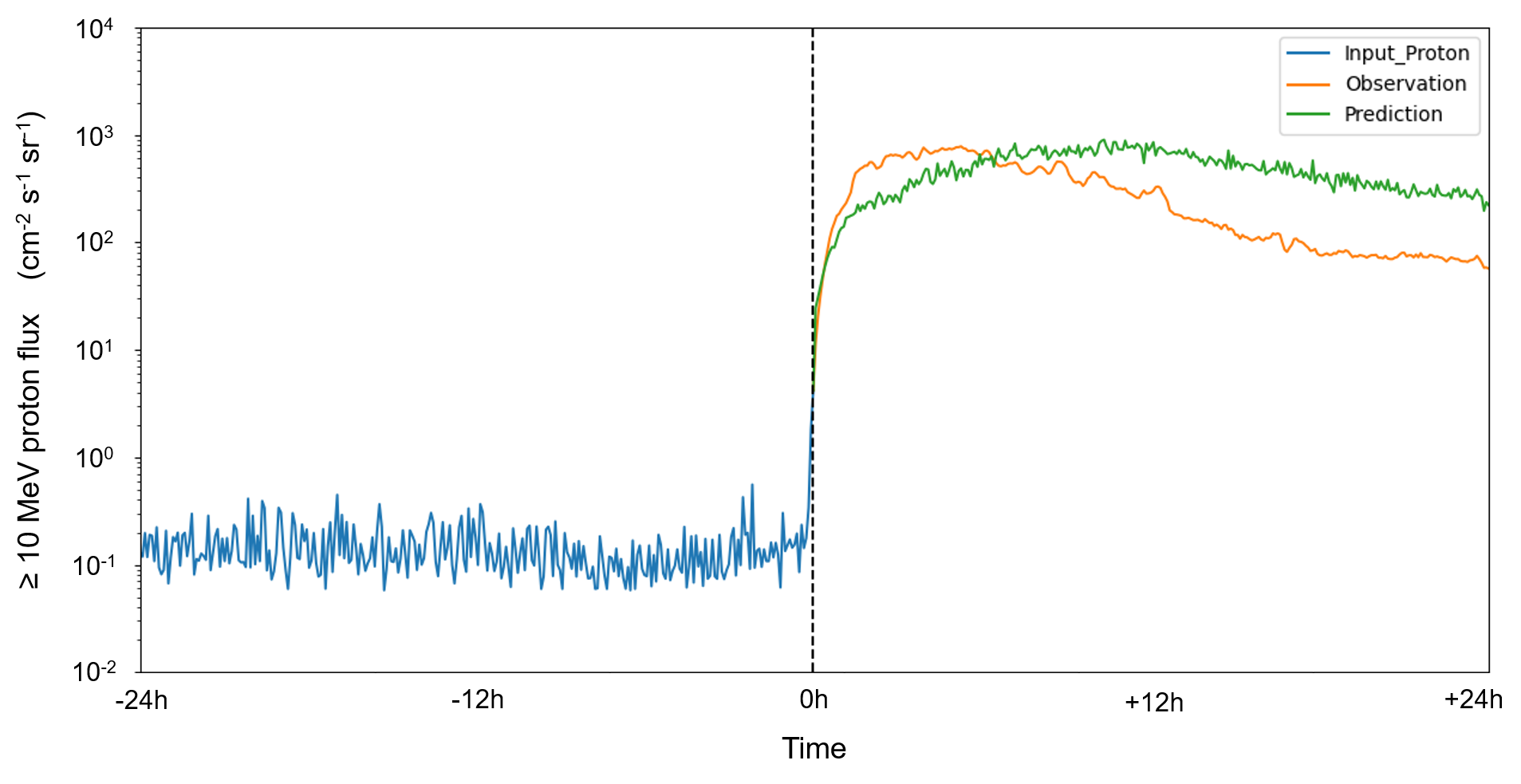}
\vspace{1mm}
\includegraphics[width=\linewidth]{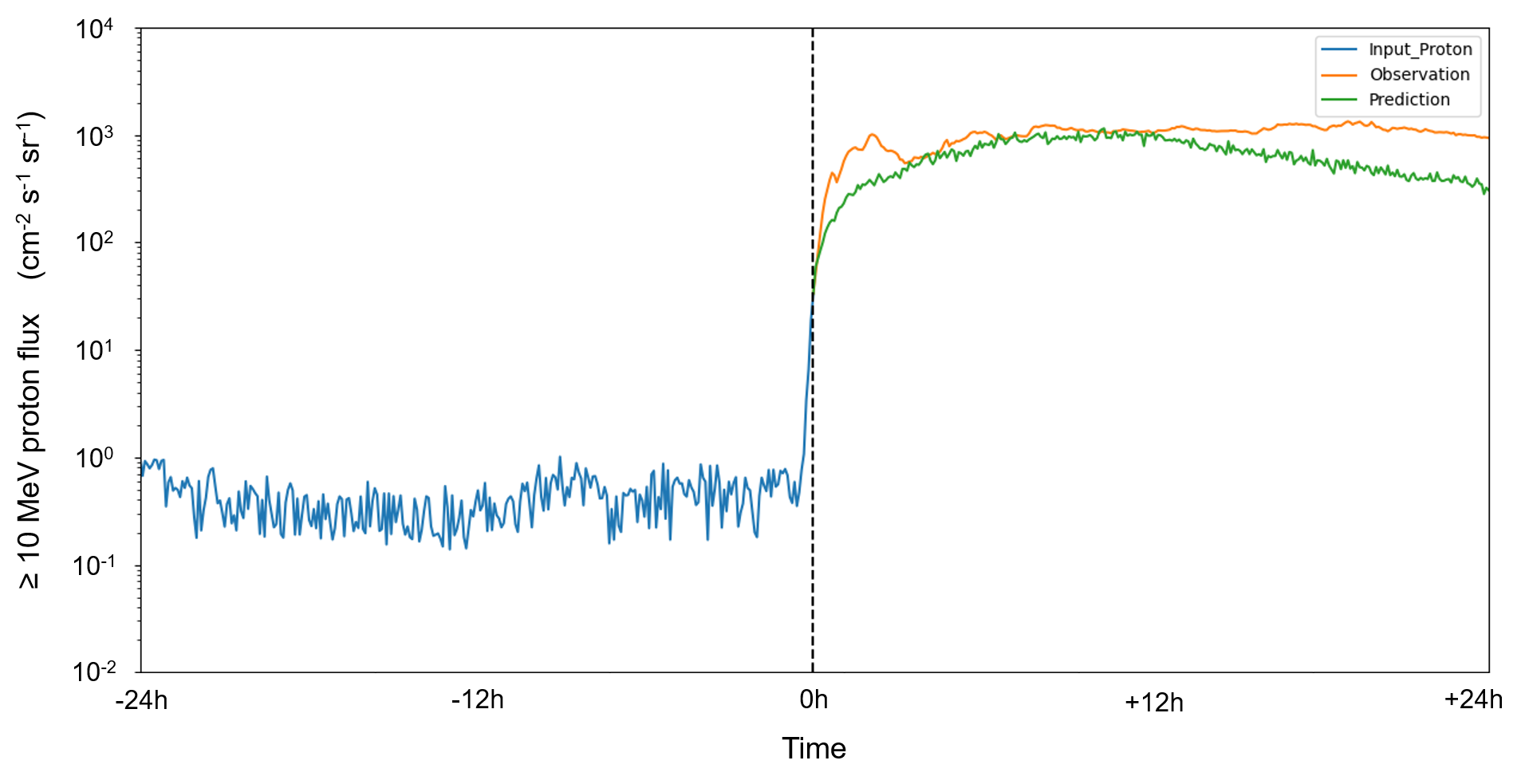}
\vspace{1mm}
\includegraphics[width=\linewidth]{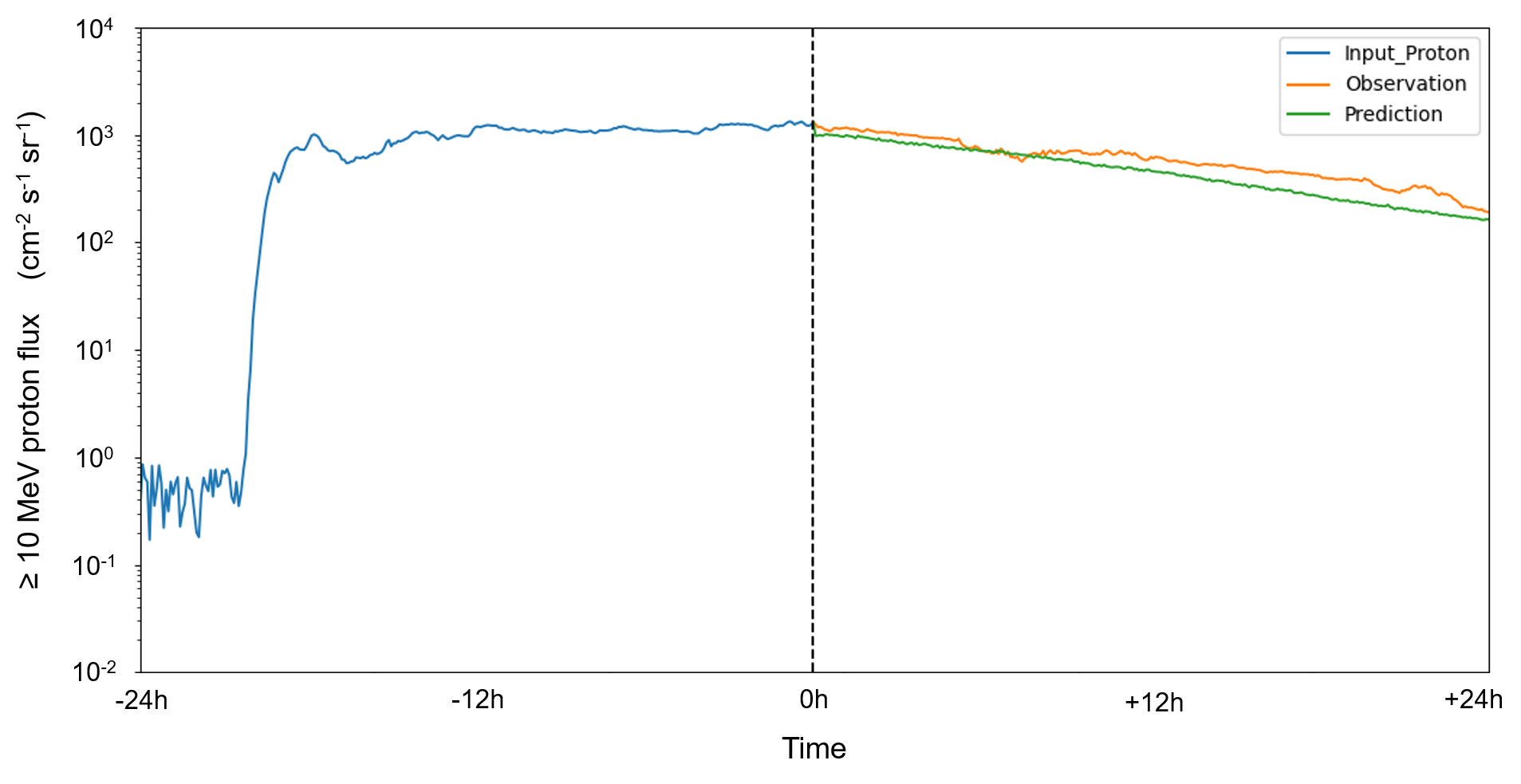}
\caption{Examples of forecasting results from the best-performance model P\_orig\_OS, 512-8 configuration). From top to bottom:
(1) S1-class SPE observed on 8 March 2011 at 01:05 UT,  
(2) S2-class SPE observed on 26 December 2001 at 06:05 UT,  
(3) S3-class SPE observed on 10 September 2017 at 16:45 UT,  
(4) Decreasing-phase region of the same S3-class SPE.  
Vertical dashed line indicates the forecasting start time. Left side of vertical line shows proton flux profiles as input, while the right side displays both the forecast and the observed values. Blue line is the input proton flux data. Orange line is the observed proton flux. Green line is the prediction result.}
\label{fig:Forecasting_example}
\end{figure}

Figure~\ref{fig:Forecasting_example} shows examples of forecasting results from the best performance model P\_orig\_OS, 512-8) for S1, S2, and S3 SPE events. The top three panels demonstrate that the model accurately predicts the peak flux and overall profile during the early phase of the SPE. The bottom panel shows that the model also performs well during the decreasing phase of the SPE. 

\section{CONCLUSION}
In this study, we investigated various deep learning strategies for forecasting the 24-hour time profile of proton flux following solar proton event (SPE) onsets. Using a dataset of 40 well-connected SPEs and a 4-fold cross-validation framework, we evaluated the performance of 15 LSTM-based seq2seq architectures under six forecasting strategies, involving different forecasting modes, input combinations, and data preprocessing techniques.

Main results are as follows.
First, one-shot forecasting generally achieves lower error than autoregressive forecasting, as it avoids error accumulation across time steps.
Second, on original data, proton-only models consistently outperform proton+X-ray models, likely due to high-frequency noise in the raw X-ray signal. However, with trend-smoothing, this gap narrows or reverses in multi-input settings, indicating improved utility of X-ray data after denoising.
Third, trend-smoothing significantly enhances performance in multi-input models by mitigating fluctuations in the X-ray channel.
Fourth, despite the overall advantage of trend-smoothed data, the best-performing model was trained on original data, suggesting that architectural choices can sometimes outweigh the benefits of data preprocessing.

This study compares forecasting strategies suited to the unique dynamics of SPE flux profiles and demonstrates the feasibility of real-time modeling using near real-time data. 
These findings offer practical insight into the design of operational forecasting systems in space weather, where low-latency inference and limited input availability are key constraints.

The small number of available events is a fundamental constraint when forecasting the flux profiles of well-connected SPEs from Earth. Comparable sample sizes have been used in prior studies, such as Ji et al. \cite{b12}, which analyzed up to 49 events. To reduce the risk of large generalization errors under this limitation, we explored various model architectures to control complexity and avoid overparameterization. We also applied 4-fold cross-validation to mitigate overfitting and obtain robust evaluation of generalization performance.

Future work will focus on expanding the data by incorporating additional real-time observations (e.g., high energy particles), generating proxy SPE profiles, and utilizing data from other spacecraft such as STEREO A and B. We also plan to explore different deep learning approaches, such as transformers and physics-informed neural networks, to improve generalization and robustness under operational conditions.

\section*{Acknowledgment}
We gratefully acknowledge the use of GOES data available online at NOAA NCEI Satellite Data Services. We also thank contributors to PyTorch, NumPy, and Matplotlib open-source packages, which were used in this study. This work was supported by NASA grant 80NSSC24M0174.


\end{document}